\documentclass[letterpaper, 10pt, conference]{ieeeconf}

\IEEEoverridecommandlockouts
\usepackage[noadjust]{cite}
\usepackage{amsmath,amssymb,amsfonts}
\usepackage{algorithmic}
\usepackage{graphicx}
\usepackage{textcomp}
\usepackage{xcolor}
\usepackage{amssymb}
\usepackage{amsmath,amssymb}
\usepackage{subcaption}
\newcommand{\cmark}{Yes}%
\newcommand{\xmark}{No}%

\usepackage{siunitx}

\usepackage{booktabs}

\usepackage{makecell}

\usepackage{tabularx}
\newcolumntype{C}{>{\centering\arraybackslash}X} 

\usepackage{ragged2e}

\usepackage[flushleft]{threeparttable}
\definecolor{tmuted}{gray}{0.6}
\newcommand{\tmuted}[0]{\textcolor{tmuted}}%

\usepackage{siunitx}

\DeclareMathOperator*{\argmin}{arg\,min}

\renewcommand{\vec}[1]{\boldsymbol{#1}}%
\renewcommand{\Vec}[1]{\boldsymbol{\mathrm{#1}}}%

\usepackage{tikz}
\usetikzlibrary{positioning}
\usetikzlibrary{fit}
\usetikzlibrary{calc}
\usetikzlibrary{shapes}
\usetikzlibrary{shapes.geometric}
\usetikzlibrary{arrows,shadows}
\usetikzlibrary{backgrounds}

\usepackage[hidelinks]{hyperref}

\usepackage{cleveref}
\crefname{figure}{fig.}{figs.}

\def\BibTeX{{\rm B\kern-.05em{\sc i\kern-.025em b}\kern-.08em
    T\kern-.1667em\lower.7ex\hbox{E}\kern-.125emX}}

\input{figures/styles.tex}

\title{Unified Perception: Efficient Depth-Aware Video Panoptic\\Segmentation with Minimal Annotation Costs}

\author{Kurt Stolle$^{1}$ \and Gijs Dubbelman$^{1}$
\thanks{* This work was supported by the NEON project which is (partly) financed by the Dutch Research Council (NWO).}
\thanks{$^{1}$ Kurt Stolle and Gijs Dubbbelman are with the Mobile Perception Systems (MPS) lab, Eindhoven University of Technology, The Netherlands.}%
}
    
\begin{document}
\thispagestyle{empty}
\pagestyle{empty}

\maketitle

\begin{abstract}
    Depth-aware video panoptic segmentation is a promising approach to camera based scene understanding.
    However, the current state-of-the-art methods require costly video annotations and use a complex training pipeline compared to their image-based equivalents.
    In this paper, we present a new approach titled Unified Perception that achieves state-of-the-art performance without requiring video-based training.
    Our method employs a simple two-stage cascaded tracking algorithm that (re)uses object embeddings computed in an image-based network.
    Experimental results on the Cityscapes-DVPS dataset demonstrate that our method achieves an overall DVPQ of 57.1, surpassing state-of-the-art methods.
    Furthermore, we show that our tracking strategies are effective for long-term object association on KITTI-STEP, achieving an STQ of 59.1 which exceeded the performance of state-of-the-art methods that employ the same backbone network.
\end{abstract}

\section{INTRODUCTION}
In this paper, we propose a unified approach to mobile perception for intelligent vehicles that combines panoptic segmentation, monocular depth estimation, and object tracking using a single, end-to-end network.
Our goal is to develop a framework that can effectively and efficiently perform these tasks in real-time, enabling the vehicle to make informed decisions and improve its overall safety and performance.
At the same time, we aim to reduce the annotation effort required for training the network, which is a major bottleneck in the development of video perception systems.

\begin{figure}[t]
    \centering
    %
\noindent{\centering\input{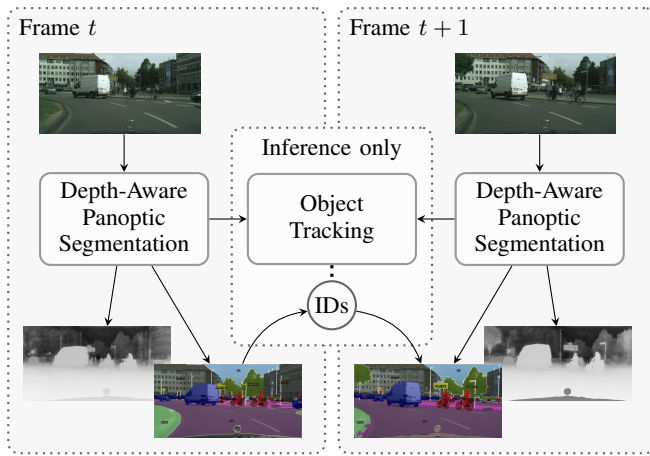}}

    \caption{\noindent{
            Overview of our proposed Unified Perception framework.
            Training is done on images alone, reducing the pipeline complexity and annotation effort required compared to video-based alternatives.
            During inference, data already computed as intermediaries in the network are (re)used to associate objects across frames.
        }
    }
    \label{fig:hero}
\end{figure}

Recent advances in deep learning have led to significant improvements in joined computer vision tasks like (depth-aware) video panoptic segmentation~\cite{Li2022VideoKNet, Qiao2020ViP, Petrovai2022,Gao2022}.
However, these methods typically rely on a tracking module that is trained on a fully annotated video sequence, which limits their generalizability to other modalities that may not have such annotations available~\cite{Weber2021STEP}. Additionally, training these methods requires significantly more labeling efforts compared to their image-based counterparts, which makes it challenging to annotate large-scale datasets~\cite{Meletis2020CityscapesParts}.
By using a tracking module that (re)uses the object embeddings that are already computed as intermediaries in an image-based (depth-aware) panoptic segmentation network, we are able to achieve state-of-the-art performance without the need for video-based training. \Cref{fig:hero} shows an overview of this concept.

Our approach is based on a state-of-the-art panoptic segmentation network, which we extend to estimate per-instance depth maps.
Because the tracker uses the intermediate object embeddings generated during image-based inference, no additional video training pipeline is required.
We experimentally validate our method on the Cityscapes-DVPS~\cite{Qiao2020ViP} dataset, outperforming state-of-the-art methods in terms of accuracy and annotation efforts.
Furthermore, we demonstrate the long-term tracking capabilities of our method on KITTI-STEP~\cite{Weber2021STEP}, where we show that the intermediate embeddings generated at each frame are sufficiently representative to follow an object over time.
Furthermore, the added inference time of our tracking module is profiled and found to be negligible, making our approach suitable for real-time applications.

The main contribution of this work is a new approach to depth-aware video panoptic segmentation that achieves state-of-the-art performance without the need for video-based training and costly annotation efforts.
Our method employs a simple tracker that reuses object embeddings computed in an image-based network, demonstrating that these intermediate data can sufficiently represent objects throughout a video for use in panoptic tracking.
The proposed approach offers a more efficient and practical solution for video based scene understanding, with potential applications in autonomous driving, robotics, and surveillance.
Our code is available at \href{https://tue-mps.github.io/unipercept}{tue-mps.github.io/unipercept}.
\section{RELATED WORK}
\subsection{Panoptic segmentation}
Panoptic segmentation ~\cite{Kirillov2018PanSeg} is a task that aims to accurately segment both object instances and stuff regions in an image, providing a comprehensive understanding of the scene.
General approaches to this task include the use of a two-stage pipeline~\cite{Kirillov2019, Liu2019E2EPanSeg, Chen2016DeepLab, Lazarow2019LearningIO}, where instance segmentation and semantic segmentation are performed independently and then fused together.
An alternative approach is the use of a single network~\cite{Li2020PanopticFCN, Sofiiuk2019AdaptISAI, Cheng2019PanDL, Chen2020ScalingWR} that is able to predict both instance-level and semantic-level segmentation maps simultaneously.
The Panoptic FCN methodology ~\cite{Li2020PanopticFCN} is a single-stage approach that has been shown to achieve state-of-the-art performance using an elegant dynamic-convolutions based approach, which has the property that each detected object has a learned compressed representation, i.e. \textit{kernel}, that could be used for tracking.

\subsection{Monocular depth estimation}
Monocular depth estimation is the task of estimating the depth map of a scene from a single image.
Historically, this task has been tackled using hand-crafted features and geometric constraints~\cite{Hoiem2005GeometricCF, Saxena2009Make3D, Liu2014DiscreteContinuousDE}.
More recently, the use of deep learning based methods has become prevalent, with many works showing that networks can be trained to predict depth maps from single images with high accuracy~\cite{Laina2016DeeperDP, Li2016ATN, AtapourAbarghouei2018RealTimeMD}.
Furthermore, the use of segmentation information in monocular depth estimation has been explored in recent works~\cite{Klingner2020SegGuidedSSMDE,Gao2022,Petrovai2022}, where depth estimates are predicted by conditioning on the instance-level features at each mask.
Panoptic Depth~\cite{Gao2022} is such a methodology that leverages instance-level segmentation masks to normalize depth estimates for every object.

In recent years, self-supervised monocular depth estimation (SSMDE) has gained significant attention as a promising approach to address the lack of noise-free ground truth depth annotations~\cite{Zhou2017SSMDE,Li2019LearningTD,Godard2019ICCV,Casser2019SSMDEEgoSem,Klingner2020SegGuidedSSMDE}.
This approach leverages the natural geometric constraints between consecutive frames in a video to train monocular depth estimation models.
Collaborative Competition~\cite{Ranjan2019Competitive} uses a depth smoothness loss for depth-aware panoptic segmentation that is based on the gradient of the estimated depth map weighted by the gradient of pixel intensities of the input image. Our method utilizes this depth smoothness loss in combination with instance-level depth normalization~\cite{Gao2022} to improve the object-level representation of depth, which yields a robust spatial cue for use in object tracking.

\subsection{Object tracking}
Object tracking entails temporally associating object detections in a video sequence, typically using spatial and appearance cues~\cite{Wang2022Unitrack, Zhang2021a,Yan2022GrandUnification,Luo2020}.
In the context of panoptic segmentation, this involves assigning a consistent and unique identifier to each indivial \textit{thing} instance that appears in a video.
Various recent methods explore this in the field of video panoptic segmentation~\cite{Kim2020VPS, Wang2021VPSSurvey}.

Annotation of object tracking data is a costly and time-consuming process, and thus, recent works have focused on developing unsupervised or weakly-supervised object tracking methods~\cite{Zhang2021, Bastani2021}.
These annotation costs are particularly high in the context of video panoptic segmentation, as in addition to consistently identifying each object, their panoptic segmentation masks must also be carefully drawn and annotated~\cite{Meletis2020CityscapesParts}.
We propose a depth-aware video panoptic segmentation network that overcomes this annotation effort by leveraging the intermediate object representations for segmentation and depth estimation to perform object tracking.
This enables us to train our network in an image-based manner, which is a significant advantage over existing methods that require ground truth annotations for full video sequences.
\section{METHODS}
\begin{figure*}[ht]
    \centering
    %
\noindent{\centering\begin{tikzpicture}[>=stealth]
    \node[text width=2mm, minimum size=0pt] (input) {$\Vec{I}$};

    \node[module, text width=12mm]
    (fpn) [right=0.5 of input]
    {FPN \cite{Kirillov2019}};

    \node[data, label={[above=1mm, anchor=south, align=center]\small Each feature}]
    (fpn-multi) [above=0.5 of fpn]
        {$\forall~\Vec{F}_i$};
    \node[circle, fill=black, minimum size=0mm, inner sep=0.5mm] (fpn-multi-out) [right=of fpn-multi] {};
    \draw[-] (fpn-multi.east) -- (fpn-multi-out);

    \node[data, label={[below=8mm, anchor=north, align=center]\small Highest-res.\\feature}]
    (fpn-single) [below=1.2 of fpn]
        {$\Vec{F}_1$};
    \node[circle, fill=black, minimum size=0mm, inner sep=0.5mm] (fpn-single-out) [right=of fpn-single] {};
    \draw[-] (fpn-single.east) -- (fpn-single-out);

    \draw[->] (input)       -- (fpn);
    \draw[->] (fpn)          -- (fpn-single);
    \draw[->] (fpn)          -- (fpn-multi);

    \node[module]           (pos)             [right=of fpn-multi-out]      {Position head};
    \node[data]                 (pos-out)         [right=of pos]            {$\Vec{L}^\mathrm{stuff}_{0 \dots N}$~,~$\Vec{L}^\mathrm{things}_{0 \dots M}$};
    \node[loss]             (pos-loss)        [right=of pos-out]        {$\mathcal{L}_\mathrm{pos}$};

    \draw[->] (fpn-multi-out)   -- (pos.west);
    \draw[-] (pos.east)     -- (pos-out);
    \draw[->] (pos-out)     -- (pos-loss);

    \node[module]           (kg-mask)     [below=0.7 of pos-out]        {Mask kernel generator};
    \draw[->] (fpn-multi-out)   |- (kg-mask.west);
    \draw[->] (pos-out)     -- (kg-mask);

    \node[module]           (kg-depth)    [above=of pos-out]        {Depth kernel generator};
    \draw[->]    (fpn-multi-out) |- (kg-depth.west);
    \draw[->]      (pos-out) -- (kg-depth);

    \node[module]           (fg-mask)    [right=of fpn-single-out]     {Mask feature generator};
    \draw[->]  (fpn-single-out) -- (fg-mask.west);

    \node[module]           (fg-depth)   [below=0.7 of fg-mask]     {Depth feature generator};
    \draw[->]  (fpn-single-out) |- (fg-depth.west);

    \node[op]               (dc-mask) [below=1 of kg-mask] {$\times$};
    \draw[->] (kg-mask)     -- node [midway, right] {$\Vec{K}_\mathrm{m}$} (dc-mask);
    \draw[->] (fg-mask)     -- node [midway, above] {$\Vec{E}_\mathrm{m}$} (dc-mask);
    \node[data]                 (out-mask) [right=2of dc-mask] {$\Vec{M}$};
    \draw[->] (dc-mask)     -- (out-mask);
    \node[loss]             (out-mask-loss) [above=of out-mask] {$\mathcal{L}_\mathrm{seg}$};
    \draw[->] (out-mask) -- (out-mask-loss);

    \node[op]               (dc-depth) [right=6.5 of fg-depth] {$\times$};
    \draw[->] (kg-depth.355) -| node [near end, right] {$\Vec{K}_d$} (dc-depth);
    \draw[->] (fg-depth) -- node [near end, above] {$\Vec{E}_\mathrm{d}$} (dc-depth);
    \node[data]                 (out-depth) [right= of dc-depth] {$\Vec{\bar{D}}$};
    \draw[->] (dc-depth) -- (out-depth);

    \node[module]           (denorm) [above=of out-depth] {IDN~\cite{Gao2022}};
    \draw[->] (out-depth)   -- (denorm);
    \draw[->] (out-mask)    -- (denorm);

    \node[data]         (depth-mr) [right=4 of kg-depth.5] {$\vec{d}_\mu~,~\vec{d}_r$};
    \draw[->] (kg-depth.5) -- (depth-mr);
    \draw[->] (depth-mr) -- (denorm);

    \node[loss] (depth-mr-loss) [right=of depth-mr] {$\mathcal{L}^\mathrm{things}_\mathrm{\mu}$};
    \draw[->] (depth-mr) -- (depth-mr-loss);

    \node[data]                 (out-depth-dn) [right = 0.5 of denorm] {$\Vec{D}$};
    \draw[->] (denorm) -- (out-depth-dn);
    \node[loss]             (out-depth-dn-loss) [above = of out-depth-dn] {$\mathcal{L}_\mathrm{depth}$};
    \draw[->] (out-depth-dn) -- (out-depth-dn-loss);
    \node[loss]                 (out-depth-loss) [below = of out-depth-dn] {$\mathcal{L}_\mathrm{sm}$};
    \draw[->] (out-depth-dn) -- (out-depth-loss);

    \node[fit=(pos)(pos-loss)(kg-mask)(fg-mask)(dc-mask)(out-mask)(out-mask-loss), region] (panoptic-fcn) {};
    \node[align=right] (panoptic-fcn-ref) [above left=0.05 of panoptic-fcn.south east] {Panoptic FCN \cite{Li2020PanopticFCN}};
\end{tikzpicture}}

    \caption{Our panoptic segmentation and depth estimation architecture.}
    \label{fig:network-architecture}
\end{figure*}
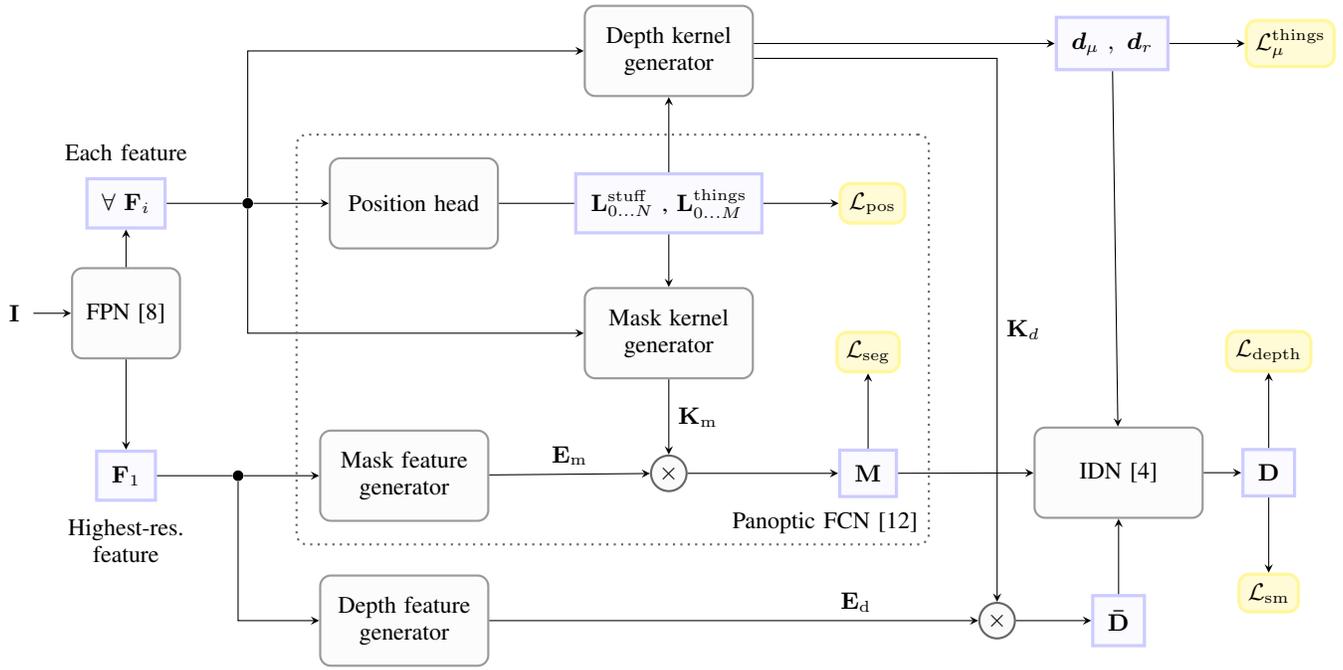

In this section, we present our framework, titled Unified Perception, which integrates the tasks of panoptic segmentation, tracking, and monocular depth estimation. To achieve this, we extend a state-of-the-art unified network architecture that incorporates dynamic convolutions for panoptic segmentation and monocular depth estimation. \Cref{fig:network-architecture} shows this architecture for image-based panoptic segmentation and monocular depth estimation.  Furthermore, by leveraging intermediate data generated by this network during inference, we introduce a tracking-by-association algorithm that effectively assigns a unique identity to each object in a video sequence by utilizing both appearance and spatial cues generated at each frame.

\subsection{Panoptic segmentation}
\label{sec:method-architecture}

We use the Panoptic FCN~\cite{Li2020PanopticFCN} framework as the basis of our network.
This architecture proposes the use of dynamic convolutions for single-path panoptic segmentation.
Specifically, the network is composed of a backbone network, a feature generator, position head, and a mask kernel generator.
The feature generator yields an embedding $\Vec{E}_m$ for each pixel in the image, and the mask kernel generator provides a mask kernel $\Vec{K}_m$ for every instance center location $\Vec{L}$ estimated by the position head.
The mask kernel is then multiplied with the embedding, i.e. dynamic convolution, to yield a segmentation mask $\Vec{M}$ for every instance.

A property of this approach is that every detected object has a corresponding kernel vector in $\Vec{K}_m$, which is trained to embed instance-level visual features.
Consequently, this kernel is a unique compressed representation of the appearance for each object in a scene.
Furthermore, we hypothesize that kernel vectors of the same object in two subsequent video frames have a degree of similarity that sufficiently distinguishes them from other objects to be used in tracking.
As will be detailed later in \Cref{sec:stage-appearance}, we leverage this property to associate objects across frames at inference-time, without the need for video-based training.

\subsection{Monocular depth estimation}
\label{sec:method-depth}

We extend the network with depth kernel and feature generator branches for monocular depth estimation, as shown in \Cref{fig:network-architecture}.
Following the paradigm of dynamic convolutions, a depth kernel embedding is generated and sampled at the same positions $\Vec{L}$ as the mask kernel embedding for every detected  instance.
This depth kernel $\Vec{K}_d$ is then multiplied with the depth embedding $\Vec{E}_d$ and followed by a sigmoid activation to yield a normalized depth map
\begin{equation}
    \Vec{\bar{D}} = \sigma(\Vec{K}_d \cdot \Vec{E}_d) \in [-1,1]
\end{equation} for every instance, which is procedurally similar to how we generate instance masks in \Cref{sec:method-architecture}. Subsequently, we implement the instance-wise depth normalization prodecure proposed in Panoptic Depth~\cite{Gao2022}. This entails offsetting the values of each normalized depth map by the mean depth $\vec{d}_\mu$ and subsequently multiply by the range $\vec{d}_r$, both of which are estimated for every instance. As discussed later in \Cref{sec:stage-spatial}, these denormalization variables provide object-level information about the location of every instance, which can be a useful cue for tracking.
As is common practice, we use the combination of scale-invariant logarithmic error and relative squared error as a loss function for the depth estimation task~\cite{Gao2022,Qiao2020ViP,Yuan2021}.

Additionally, we note that ground truth depth maps often contain noise and generation artifacts.
As experimentally validated in Mono DVPS~\cite{Petrovai2022}, a self-supervized smoothness loss can make the depth estimator more robust to such ground truth noise.
Based on the methodology proposed in~\cite{Ranjan2019Competitive}, we define a pixel-intensity weighted smoothness loss for each instance-level depth map $\Vec{\bar{D}}$ as
\begin{equation}
    \mathcal{L}_\mathrm{sm} =
    \left| \partial_x \Vec{\bar{D}} \right|
    e^{ -\left|| \partial_x \Vec{I} \right|| }
    +
    \left| \partial_y \Vec{\bar{D}} \right|
    e^{ -\left|| \partial_y \Vec{I} \right||}
\end{equation}
where $x$ and $y$ are the horizontal and vertical directions, respectively, and $\Vec{I}$ is the input image converted to grayscale.

\subsection{Object tracking}

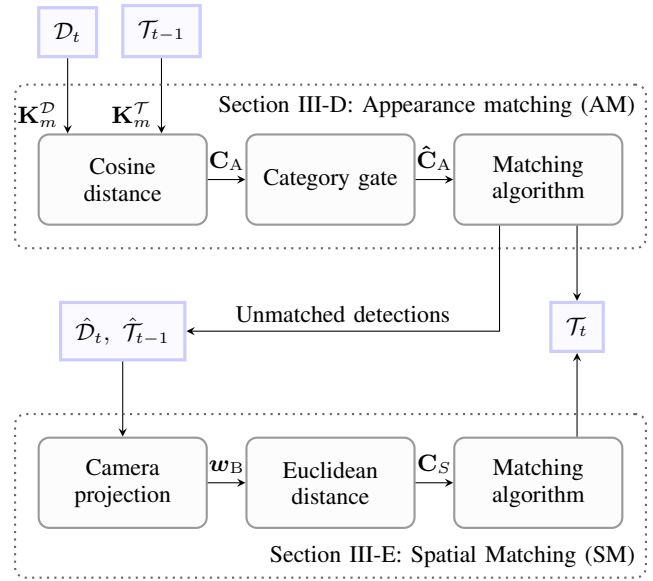
\begin{figure}
    \centering
    %
\noindent{\centering\begin{tikzpicture}[>=stealth]

    \node[module]   (cost-am)                      {Cosine distance};
    \node[] (stage-am-in) [above=0.1 of cost-am] {};
    \node[data] (D_in)  [above=1 of cost-am.140]      {$\mathcal{D}_t$};
    \node[data] (T_in)  [above=1 of cost-am.50]      {$\mathcal{T}_{t-1}$};
    \draw[->] (D_in) -- node [near end, left] {$\Vec{K}_m^\mathcal{D}$} (cost-am.140);
    \draw[->] (T_in) -- node [near end, left] {$\Vec{K}_m^\mathcal{T}$} (cost-am.50);

    \node[module]   (gate-am)  [right=0.5 of cost-am]     {Category gate};
    \draw[->] (cost-am) -- node [midway, above] {$\Vec{C}_\mathrm{A}$} (gate-am);

    \node[module]   (asgn-am)  [right=0.5 of gate-am]     {Matching algorithm};
    \draw[->] (gate-am) -- node [midway, above] {$\Vec{\hat{C}}_\mathrm{A}$} (asgn-am);

    \node[data]         (match) [below=of asgn-am.310]     {$\mathcal{T}_t$};
    \draw[->] (asgn-am.310) -- (match);
    \draw[->, dotted, thick] (match);

    \node[fit=(stage-am-in)(asgn-am)(gate-am)(cost-am)(cost-am), region] (stage-am) {};
    \node[align=right] (stage-am-ref) [below left=0.05 of stage-am.north east] {\Cref{sec:stage-appearance}: Appearance matching (AM)};

    \node[data]         (unmatched)  [below=1 of cost-am]      {$\hat{\mathcal{D}}_t,~\hat{\mathcal{T}}_{t-1}$};
    \draw[->] (asgn-am.230) |- node [near end, above] {Unmatched detections} (unmatched);
    \node[module]   (cost-sm) [below=1 of unmatched]                  {Camera projection};
    \node[] (stage-sm-in) [below=0.1 of cost-sm] {};

    \draw[->] (unmatched) -- (cost-sm);

    \node[module]   (gate-sm)  [right=0.5 of cost-sm]     {Euclidean distance};
    \draw[->] (cost-sm) -- node [midway, above] {$\vec{w}_\mathrm{B}$} (gate-sm);

    \node[module]   (asgn-sm)  [right=0.5 of gate-sm]     {Matching algorithm};
    \draw[->] (gate-sm) -- node [midway, above] {$\Vec{C}_S$} (asgn-sm);
    \draw[->] (asgn-sm.50) -- (match);

    \node[fit=(stage-sm-in)(asgn-sm)(gate-sm)(cost-sm)(cost-sm), region] (stage-sm) {};
    \node[align=right] (stage-sm-ref) [above left=0.05 of stage-sm.south east] {\Cref{sec:stage-spatial}: Spatial Matching (SM)};

\end{tikzpicture}}

    \caption{Our proposed tracking architecture using a two-stage assignment algorithm that makes use of appearance and spatial cues, where $\mathcal{D}$ is the set of detections and $\mathcal{T}$ is the set of tracklets.}
    \label{fig:tracker-modules}
\end{figure}

At every frame, the \textit{tracker} module is tasked with assigning a temporally consistent track identifier to every unique \textit{thing} instance that appears in a video sequence.
The tracker is applied at every subsequent frame (online), and thus has no access to information from future frames.
The structure that defines a detection $\delta \in \mathcal{D}$ proposed by the network at a frame is a tuple of intemediate data from the network, i.e.
\begin{equation}
    \mathcal{\delta} = (\Vec{L}, \Vec{K}_m, \vec{d}_\mu).
\end{equation}
When the tracker assigns an identifier $\iota \in \mathbb{N}$ to a detection, the resulting structure is canonically named a tracklet $\tau \in \mathcal{T}$, and is represented by
\begin{equation}
    \tau = (\delta, \iota)
\end{equation}
We feed detections through a sequence of tracking-by-association stages.
Every stage performs a matching procedure based on a linear sum assignment formulation.
Specifically, we compute a cost function
\begin{equation}
    \Vec{C}
    :
    (\mathcal{D},\mathcal{T})
    \longrightarrow
    \mathbb{R}^{|\mathcal{D}|\times|\mathcal{T}|}_+
\end{equation}
that represents the assignment cost from all detections $\mathcal{D}_t$ at the current time-step $t$ to tracklets $\mathcal{T}_{t-1}$ at the previous frame.
Next, we formulate a linear assignment problem to yield a matching
\begin{equation}
    \mathcal{M} \subset \{(\delta_i, \tau_n)\cdots(\delta_j, \tau_m)~|~i \ne j, n \ne m \}
\end{equation}
such that
\begin{equation}
    \mathcal{M}^\star = \argmin_{\mathcal{M}} \sum_{
        m \in \mathcal{M}
    }
    \Vec{C}(m)
\end{equation}
where $\mathcal{M}^\star$ represents the optimal combination of detections and tracklets that results in the smallest total assignment cost. We apply the Jonker-Volgenant algorithm~\cite{Jonker1987LAP} to solve this formulation.

The identifier $\iota$ of each matched tracklet is propagated to the corresponding detection, and unmatched detections and tracklets are passed on to the next tracker stage. Finally, any remaining unmatched detections are assigned to newly generated identifiers.

\Cref{fig:tracker-modules} shows a diagram of our proposed tracker.
By using two cascaded stages that respectively use a cost function over appearance cues from the mask kernel vector $\Vec{K}_m$ (\Cref{sec:stage-appearance}) and spatial cues from the mean instance depth $\Vec{d}_\mu$ (\Cref{sec:stage-spatial}), we aim to define a tracking pipeline that is robust to noise that arises due to not enforcing consistency for each (re)used object representation through video-based training, compared to single-stage tracking with either appearance or spatial representations.

\subsection{Appearance matching (AM) stage}
\label{sec:stage-appearance}
The appearance matching stage enables the effective association of objects across consecutive frames based on their visual features.
In our case, the mask kernel vector $\Vec{K}_m$ is a compact representation that encodes the visual features of the object.
We define the cost function via the cosine similarity, which calculates the angle between two vectors, from the mask kernel vectors of detections $\mathcal{D}$ and tracklets $\mathcal{T}$.
The cost function is thus defined as
\begin{equation}
    \Vec{C}_\mathrm{A}(\Vec{K}_m^\mathcal{D}, \Vec{K}_m^\mathcal{T})
    = 1 -
    \frac{
        \Vec{K}_m^\mathcal{D} \cdot \Vec{K}_m^\mathcal{T}
    }{
        ||\Vec{K}_m^\mathcal{D}||~||\Vec{K}_m^\mathcal{T}||
    }
\end{equation}
where the similarity is subtracted from one to yield a minimization objective.
As is common practice~\cite{Zhang2021}, we apply a gategory gate to the cost function, where entries that have a different predicted class label are set to have infinite cost.

\subsection{Spatial matching (SM) stage}
\label{sec:stage-spatial}
Given that the mean intance depths $\vec{d}_\mu$ include an estimate of the mean depth of each object, we aim to define a cost function that utilizes this intermediate data as a proxy for the similarity between the positions of objects in the previous to the current frame.
To do this, we make two assumptions about the movement of the camera and the objects that we aim to track.
First, we assume that the camera is placed on an Ackermann drive, which is a type of steering mechanism commonly used in vehicles. This implies that the camera is not capable of pure rotation.
Second, we assume that objects move predominantly in the horizontal plane, which is a valid assumption for most intelligent vehicle applications.
Based on these assumptions, we propose creating a 2D \textit{birds-eye-view} representation of objects in the scene, where object coordinates are mapped to points on a horizontal plane that is parallel to the ground.
We project each object to this plane using the camera intrinsics, mean depth $\vec{d}_\mu$, and object center coordinates in the image $\Vec{L}$.
As the spatial cost function we use the Euclidean distance, which represents the similarity between two positions in the scene, i.e.
\begin{equation}
    \Vec{C}_\mathrm{S}(\vec{w}_\mathrm{B}^\mathcal{D}, \vec{w}_\mathrm{B}^\mathcal{T}) = ||\vec{w}_\mathrm{B}^\mathcal{D} - \vec{w}_\mathrm{B}^\mathcal{T} ||
\end{equation}
where $\vec{w}_\mathrm{B}$ represents an object coordinate in the birds-eye-view.
\section{EXPERIMENTS}

In this section, we experimentally assess the performance of our proposed method.
First, we employ the depth-aware panoptic segmentation task on Cityscapes-DVPS~\cite{Qiao2020ViP}. This shows the overall performance of the network when all target tasks are considered: panoptic segmentation, monocular depth estimation and object tracking.
Second, we focus on panoptic tracking using the KITTI-STEP dataset~\cite{Weber2021STEP}.
Here, we investigate the long-term associative capabilities of the network under ablation of our proposed tracking stages.

\begin{figure*}
    \centering
    \input{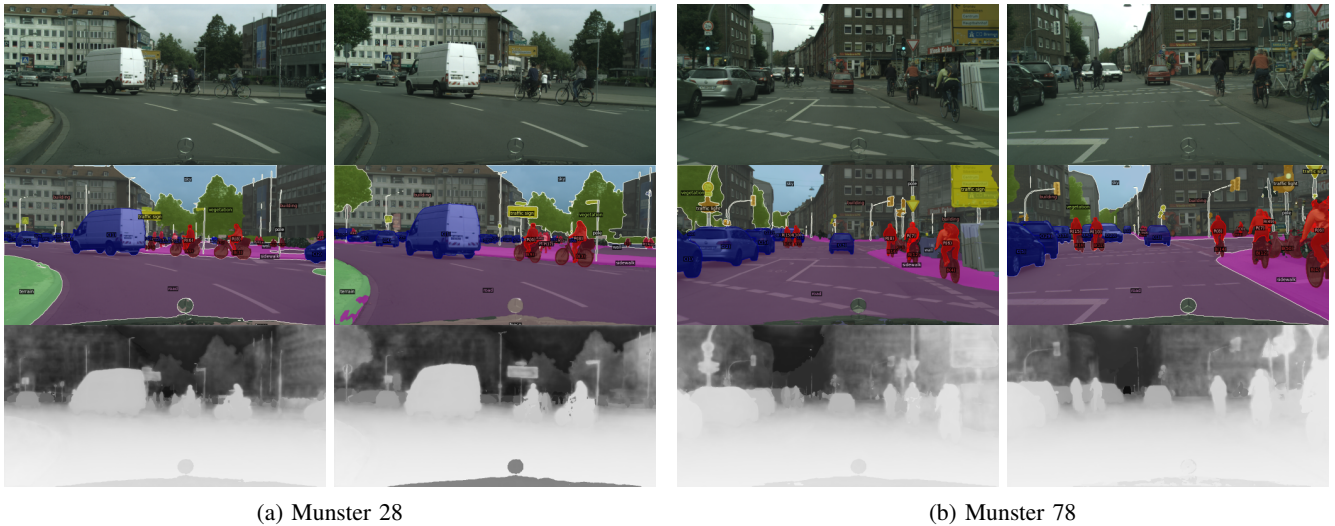}
    \caption{Qualitative results on the Cityscapes-DVPS dataset.}
    \label{fig:qualitative-cs-dvps}
\end{figure*}

\begin{table*}
    \centering
    \begin{threeparttable}

        \caption{Results on Cityscapes-DVPS dataset under ablation of depth normalization and smoothness loss}
        \label{tab:results-cs-dvps}
        \begin{tabularx}{\linewidth}{X c c c c c c c}
            \toprule

            \thead[l]{ Method}                           &
            \thead{Instance depth                                              \\normaliztion\tnote{a}}            &
            \thead{Depth                                                       \\
            smooth loss\tnote{b}}                        &
            \thead{Backbone                                                    \\network}                             &
            \thead{Video training                                              \\
            required}                                    &
            \thead{DVPQ $\uparrow$                    }  &
            \thead{DVPQ\textsuperscript{th} $\uparrow$ } &
            \thead{DVPQ\textsuperscript{st} $\uparrow$        }                \\
            \midrule
            Unified Perception (\textit{ours})
                                                         & \xmark
                                                         & \xmark
                                                         & ResNet-50
                                                         & \xmark
                                                         & 43.9
                                                         & 33.2
                                                         & 50.5                \\
            Unified Perception (\textit{ours})
                                                         & \cmark
                                                         & \xmark
                                                         & ResNet-50
                                                         & \xmark
                                                         & 50.4
                                                         & 44.3
                                                         & 53.4                \\
            Unified Perception (\textit{ours})
                                                         & \xmark
                                                         & \cmark
                                                         & ResNet-50
                                                         & \xmark
                                                         & 45.4
                                                         & 36.0
                                                         & 51.4                \\
            Unified Perception (\textit{ours})
                                                         & \cmark
                                                         & \cmark
                                                         & ResNet-50
                                                         & \xmark
                                                         & \textbf{57.1}
                                                         & \textbf{57.4}
                                                         & 57.6                \\
            \midrule

            PolyphonicFormer~\cite{Yuan2021}             & ~             & ~
                                                         &
            ResNet-50
                                                         &
            \cmark
                                                         &
            48.1
                                                         &
            35.6
                                                         &
            57.1                                                               \\
            Mono DVPS~\cite{Petrovai2022}                & ~             & ~ &
            ResNet-50                                    &
            \cmark                                       &
            50.4                                         &
            35.9                                         &
            \textbf{61.7}                                                      \\
            \midrule
            \tmuted{ViP-DeepLab~\cite{Qiao2020ViP}}
                                                         & ~             & ~ &
            \tmuted{WR-41}                               &
            \tmuted{\cmark}                              &
            \tmuted{ 55.1}                               &
            \tmuted{43.3}                                &
            \tmuted{\bfseries 63.6}                                            \\
            \bottomrule
        \end{tabularx}

        \smallskip
        \scriptsize
        \begin{tablenotes}
            \RaggedRight
            \item[a] Instance depth normalization~\cite{Gao2022} enabled, otherwise all depth maps are normalized to the same scale.
            \item[b] Edge-aware depth smoothness loss~\cite{Ranjan2019Competitive} enabled during all training and finetuning.
        \end{tablenotes}

    \end{threeparttable}
\end{table*}

\subsection{Datasets}
\textbf{Cityscapes-DVPS:}
We use the Cityscapes-DVPS benchmark proposed in ViP-DeepLab~\cite{Qiao2020ViP}, which is an extension of Cityscapes-VPS~\cite{Kim2020VPS} with added depth maps computed using stereo disparity.
Each sequence in this dataset consists of 30 frames, where every 5th frame is annotated with a panoptic segmentation and depth map.
Following the canonical evaluation procedure, results are interpreted using the Depth-aware Video Panoptic Quality (DVPQ) metric.
This metric essentially measures the video panoptic quality~\cite{Kim2020VPS} when only pixels are considered for which the absolute relative depth error is below a threshold value.

\textbf{KITTI-STEP:}
We aim to evaluate the long-term tracking capabilities of our methodology using the KITTI-STEP dataset \cite{Weber2021STEP}, which is evaluated using the Segmentation and Tracking Quality (STQ).
This metric is defined as the geometric mean between the association quality (AQ), which measures the tracking performance, and the segmentation quality (SQ).
This separation of association and segmentation scores is important, as it allows us to evaluate the tracking performance of our method independently of the panoptic segmentation performance.
The dataset uses the Cityscapes labeling scheme, and training can thus be initialized from the trained weights of the Cityscapes-DVPS model.
Furthermore, because the KITTI-STEP dataset does not provide ground-truth depth maps, we use only the depth smoothness loss (\Cref{sec:method-depth}) to train our monocular depth estimation modules.
While this is not ideal for the task of depth estimation, we find that it is sufficient to yield an object-level depth estimate that is useful for tracking.

\subsection{Implementation details}
We use ResNet-50~\cite{He2015ResNet} as the backbone network, and pre-train its weights  on COCO~\cite{Lin2015COCO}.
Our network is then trained on Cityscapes~\cite{Cordts2016} for 80K steps, following the same training settings from Panoptic FCN~\cite{Li2020PanopticFCN}.
Ground-truth depth maps are generated using the stereo-disparity maps provided in the dataset supplementary materials.
We note that these depths are likely computed using different methods than the proprietary algorithm used to generate those provided in Cityscapes-DVPS~\cite{Qiao2020ViP}, resulting in a small domain gap between the two datasets.
However, we emperically find that the addition of the depth-smoothness loss (\Cref{sec:method-depth}) appears to effectively regularize our network towards a generalized representation that is not negatively impacted by this change in modalities.
Next, the model is finetuned on a shuffled set of frames from the Cityscapes-DVPS~\cite{Qiao2020ViP} sequences for 20K steps using the AdamW optimizer and a learning rate of \num{1e-8}.
Finally, for the KITTI-STEP~\cite{Weber2021STEP} dataset, we finetune for an additional 10K steps with the same hyperparameters.

\subsection{Main results}

\begin{table*}
    \centering
    \begin{threeparttable}

        \caption{Results on KITTI-STEP dataset under ablation of tracker modules}

        \label{tab:results-kitti-step}

        \begin{tabularx}{\linewidth}{X c c c c c c c c c}
            \toprule

            \thead[l]{Method}                                       &
            \thead{Tracker                                            \\stages\tnote{a}}                  &
            \thead{Backbone                                           \\network}                                 &
            \thead{Video training                                     \\required}                  &
            \thead{STQ $\uparrow$                        }          &
            \thead{AQ $\uparrow$                         }          &
            \thead{SQ  $\uparrow$                        }            \\

            \midrule

            Unified Perception (\textit{ours})                      &
            AM                                                      &
            ResNet-50                                               &
            \xmark                                                  &
            56.8                                                    &
            51.9                                                    &
            61.9                                                      \\
            Unified Perception (\textit{ours})                      &
            SM                                                      &
            ResNet-50                                               &
            \xmark                                                  &
            53.8                                                    &
            46.9                                                    &
            61.9                                                      \\
            Unified Perception (\textit{ours})                      &
            AM$\rightarrow$SM                                       &
            ResNet-50                                               &
            \xmark                                                  &
            \textbf{ 59.1 }                                         &
            \textbf{ 56.4 }                                         &
            \textbf{ 61.9 }                                           \\

            \midrule

            Motion-DeepLab~\cite{Weber2021STEP}                     &
            ~                                                       &
            ResNet-50                                               &
            \cmark                                                  &
            52.2                                                    &
            45.6                                                    &
            59.8                                                      \\

            \midrule

            \tmuted{SIAin~\cite{Ryu2021SIAin}}                      &
            ~                                                       &
            \tmuted{Swin-L}                                         &
            \tmuted{\cmark}                                         &
            \tmuted{57.9}                                           &
            \tmuted{55.2}                                           &
            \tmuted{60.7}                                             \\

            \tmuted{TubeFormer-DeepLab~\cite{Kim2022TubeFormer-DL}} &
            ~                                                       &
            \tmuted{Axial-ResNet-50}                                &
            \tmuted{\cmark}                                         &
            \tmuted{\bfseries 65.2}                                 &
            \tmuted{\bfseries 61.0}                                 &
            \tmuted{\bfseries 70.3}                                   \\

            \bottomrule
        \end{tabularx}

        \smallskip
        \scriptsize
        \begin{tablenotes}
            \RaggedRight
            \item[a] Stages enabled in our tracking algorithm, see \Cref{fig:tracker-modules}
        \end{tablenotes}

    \end{threeparttable}
\end{table*}

In \Cref{tab:results-cs-dvps} we compare our proposed framework to the state-of-the-art on Cityscapes-DVPS~\cite{Qiao2020ViP}.
In our interpretation, we only compare against models that are trained using the same backbone network, as we find that the use of a larger backbone network results in an unfair comparison.
Our approach achieves significantly improved performance compared to the evaluated methodologies.
This is largely due to an improved score for \textit{thing}-classes, which we observe to be attributable to the use of instance depth normalization proposed by Panoptic Depth~\cite{Gao2022}.
Additionally, experimental results show that use of the depth smoothness loss~\cite{Ranjan2019Competitive} alone does not yield competitive performance.
However, while not strictly required to achieve competitive performance, using this in combination with depth normalization significantly improves the performance of our model beyond that of existing approaches.
\Cref{fig:qualitative-cs-dvps} shows a qualitative visualization.


Results on KITTI-STEP~\cite{Weber2021STEP} are summarized in \Cref{tab:results-kitti-step}, under ablation of each tracker stage.
We find that the full cascade of AM and SM stages significantly outperforms the baseline network Motion-DeepLab~\cite{Weber2021STEP}, having an STQ of 59.1.
Furthermore, albeit our depth maps may be suboptimal due to the lack of a ground truth during training, the mean depths have significant value in tracking as shown by the attained AQ score of 46.9 using only the SM stage, which surpasses the baseline AQ of 45.6.
This is likely due to that the mean depth is a good proxy for the distance to the camera, which is a key factor in tracking.

\begin{table}[t]
    \centering
    \begin{threeparttable}

        \caption{Inference time $T$ in milliseconds on GPU}
        \label{tab:results-time}

        \begin{tabularx}{\linewidth}{X C C C C}
            \toprule

            ~                                            &
            \multicolumn{2}{c}{\textbf{Cityscapes-DVPS}} &
            \multicolumn{2}{c}{\textbf{KITTI-STEP}}               \\

            \cmidrule{2-3}
            \cmidrule{4-5}
                                                         &
            $T$                                          &
            $\Delta$\tnote{b}                            &
            $T$                                          &
            $\Delta$\tnote{b}                                     \\

            \midrule

            Baseline\tnote{a}                            &
            154                                          & {--} &
            70                                           & {--}   \\

            \midrule

            AM                                           &
            160                                          & 6    &
            73                                           & 3      \\

            SM                                           &
            164                                          & 10   &
            76                                           & 6      \\

            AM$\rightarrow$SM                            &
            165                                          & 11   &
            77                                           & 7      \\

            \bottomrule
        \end{tabularx}

        \smallskip
        \scriptsize
        \begin{tablenotes}
            \RaggedRight
            \item[a] All tracking modules disabled. Object identifiers are randomly assigned.
            \item[b] Difference with respect to baseline.
        \end{tablenotes}

    \end{threeparttable}
\end{table}

Finally, we investigate the deployment aspects of our methodology by providing an ablation on the effects of each tracking stage on inference time.
We profiled the mean inference time over 100K frames on a single NVIDIA RTX A6000 GPU with a batch size of one. As a baseline, we profiled the network when no tracking algorithm is used.
It should be noted that this configuration does not yield functional results, but does enable computing the added overhead for each stage.
\Cref{tab:results-time} summarizes the results of this experiment.
Our full two-stage tracker adds \num{11} ms and \num{7} ms to the inference time of the panoptic segmentation and monocular depth estimation model on respectively Cityscapes-DVPS and KITTI-STEP.
This is an acceptable overhead for the significant performance gains that our approach provides.
Furthermore, we have not yet optimized our model for production use.
We believe that such efforts would further reduce the inference time, enabling real-time inference on modern GPU-based embedded systems.

\section{CONCLUSION}
We proposed Unified Perception, a framework for joint panoptic segmentation, monucular depth estimation and object tracking. Because our model was trained on images alone, the annotation costs involved in deploying this methodology are drastically lower compared to state-of-the-art alternatives that learn an object representation over a video sequence.

In our approach, we hypothesized that the intermediate object representations generated in an image-based network for depth-aware panoptic segmentation provide enough information to facilitate tracking.
Our results indeed show that the proposed methods can achieve significantly improved performance on Cityscapes-DVPS and KITTI-STEP over alternative video-based methods.
We note that the performance could potentially be further improved when using a larger and more expressive backbone network, such as the Swin Transformer~\cite{Liu2021} or Axial-ResNet~\cite{Wang2020Axial}, which we leave for future work.

While other depth-aware video panoptic segmentation methods provide valuable insights into novel ways to model scene understanding, our approach shows that the increased complexity from video-based training is not a hard requirement for attaining competitive results.
We hope that this research inspires new ways to leverage the (re)use of intermediate object descriptors for miscellaneous tasks, leading to more conservative annotation costs.

\bibliographystyle{IEEEtran}
\bibliography{library}

\end{document}